# TempNet – Temporal Super Resolution of Radar Rainfall Products with Residual CNNs

Muhammed Sit[a*], Bong-Chul Seo[a], Ibrahim Demir [a,b,c]

[a] IIHR Hydroscience and Engineering, University of Iowa, Iowa City, Iowa, USA
[b] Civil and Environmental Engineering, University of Iowa, Iowa City, Iowa, USA
[c] Electrical and Computer Engineering, University of Iowa, Iowa City, Iowa, USA

* Corresponding Author, Email: muhammed-sit@uiowa.edu

**Abstract**

The temporal and spatial resolution of rainfall data is crucial for environmental modeling studies in which its variability in space and time is considered as a primary factor. Rainfall products from different remote sensing instruments (e.g., radar, satellite) have different space-time resolutions because of the differences in their sensing capabilities and post-processing methods. In this study, we developed a deep learning approach that augments rainfall data with increased time resolutions to complement relatively lower resolution products. We propose a neural network architecture based on Convolutional Neural Networks (CNNs) to improve the temporal resolution of radar-based rainfall products and compare the proposed model with an optical flow-based interpolation method and CNN-baseline model. The methodology presented in this study could be used for enhancing rainfall maps with better temporal resolution and imputation of missing frames in sequences of 2D rainfall maps to support hydrological and flood forecasting studies.

## 1. Introduction

From disaster preparedness to response and recovery needs, the availability and quality of environmental datasets have gained significant importance in recent years with the increased impact of natural disasters. Rainfall datasets have been an important component in many modeling applications such as flood forecasting (Seo et al., 2018, 2021; Sit et al., 2021a; Xiang and Demir, 2022a), water quality modeling (Jha et al., 2007; Demir et al., 2009), wastewater management (Cahoon and Hanke, 2017), along with other data products based on sensor networks and instrumentation (Muste et al., 2017). Since spatial and temporal distributions of rain exert importance in such modeling efforts, the quality and availability of precipitation maps hold the utmost importance in advancing research on disaster mitigation (Teague et al., 2021; Alabbad et al., 2022) and risk assessment (Yildirim and Demir, 2021), and decision support (Ewing and Demir, 2021).

Rainfall products from Quantitative Precipitation Estimation (QPE) systems are three-dimensional, including the time component. The first two dimensions reflect spatial coordinates on earth, including latitude and longitude. The third dimension here is temporal resolution. Weather radars have been a primary instrument for QPE and allow us to capture space-time features of rainfall, which is required for environmental (i.e., hydrologic) predictions at relevant space-time scales. Due to its sensing nature, the accuracy of QPE broadly depends on many factors (Villarini and Krajewski 2010). Since a composite of multiple radars is often used to produce large-scale rain maps, computational capabilities and methodology play an essential role in the process of combining multiple radar data. Consequently, it should be noted that synchronization of different observation times among multiple radars (interpolation or extrapolation over time) is a major challenge to generate a composite rain map. Another QPE issue regarding space-time resolutions of radar product is the misrepresentation of rainfall accumulation arising from radar's intermittent sampling strategy (e.g., Fabry et al., 1994; Liu and Krajewski, 1996; Seo and Krajewski, 2015).

Mitigating the uncertainty caused by radars in rainfall products has been an essential task in radar hydrology (Krajewski and Smith, 2002). Most efforts in radar hydrology for better rainfall datasets rely on understanding uncertainty factors and developing their technical solution in the processing algorithms. Regarding the technology used, efforts usually focus on getting more accurate precipitation data for a better sense of weather in terms of space or time. Once the data has been acquired, however, the effort of generating better datasets becomes a different task.

The temporal resolution of rainfall products is one of the key elements that determine the prediction accuracy of modeling efforts (e.g., Atencia et al., 2011). Interpolating 2D rain maps is not a unique problem for rainfall estimation. The same approach has been a topic of interest in the field of computer vision and, with advancements in computational capacities, in the deep learning applications field, precisely for video frame interpolation, or in other words, video temporal interpolation. Recording every video at high frame rates is impractical or expensive for many settings. Thus, developing models or methods to increase video frame rates has been extensively studied in recent years. In computer vision, temporal interpolation is done with both conventional statistical methods and data-driven machine learning models. Even though proposed solutions could be realized on 1-channel versions of 2D images that form videos, performing temporal interpolations over 3-channel images is more explored, making the task at hand a problem of estimating the tensor between two 3D tensors. Regardless, the methodology that could be employed and challenges remain vastly similar among video temporal interpolation and temporal interpolation of 2D rain maps.

With inspiration from video frame interpolation studies, this paper proposes a convolutional neural networks (CNNs) based deep learning architecture to account for radar's intermittent sampling and thus improve its product temporal resolution. The proposed CNN model, Temporal Resolution Network, which will be referred to as TempNet throughout this paper, is compared to three baseline methods: the nearest frame, optical flow and a relatively simple but deep CNN that will be referred to as CNN-baseline. As with baseline methods to compare, the proposed CNN, TempNet produces an intermediate 2D rain map between two temporally consecutive rain maps. In other words, the methodology presented here aims to perform temporal interpolation of 2D rain maps.

The rest of this paper is structured as follows: in section 2, a brief review of the literature is provided. Subsequently, in section 3, the overview of the methodology is presented, starting with the description of the dataset used, followed by the proposed deep neural networks and baseline methods to compare them. In section 4, the results from the nearest frame, optical flow, CNN-baseline, and the TempNet are presented and discussed. Finally, in section 5, the conclusions are summarized with final remarks regarding the study.

## 2. Related Work

Research studies for better rainfall products can be broadly classified as either quality- or resolution-related. Dataset improvements using deep neural networks include data cleaning to eliminate noise (Lepetit et al., 2021), increasing the resolution or accuracy of datasets by various statistical or data-driven methods (Li et al., 2019; Demiray et al., 2021a; Demiray et al., 2021b), synthetic data generation (Gautam et al., 2020), and bias correction (Hu et al., 2021). Resolution-related improvements, on the other hand, either focus on increasing the resolution of two spatial dimensions or the temporal dimension.

Unlike deep learning literature, in meteorology and climatology, whether it be temporal or spatial, super-resolution is referred to as downscaling. Downscaling of precipitation products has been a topic of interest for domain scientists. For instance, Hou et al. (2017) built an ensemble of a Markov chain and a Support Vector Machine to increase the spatial resolution of daily precipitation. Likewise, Vandal et al. (2019) explored five statistical downscaling methods for Global Climate Models (GCMs) including machine learning approaches such as autoencoders and support vector machines. Beyond traditional machine learning models, ANNs have been employed for spatial downscaling as well. Alizamir et al. (2017) integrated Particle Swarm Optimization (PSO) into ANNs for arid regions. Similarly, Salimi (2019) et al., utilized PSO, ANNs, as well as genetic algorithm for downscaling of precipitation events in Austria.

Utilizing adversarial training, Chaudhuri and Robertson (2020) proposed CliGAN, a Generative Adversarial Network (GAN) that uses an encoder-decoder structure for the generator, trained with a combination of adversarial loss, content loss with the Nash-Sutcliff Model Efficiency (NSE) and structural loss with the multi-scale structural similarity index (MSSIM). The authors showed that CliGAN can downscale an ensemble of large-scale annual global maximum precipitation products to regional precipitation products. CNNs were used by Tu et al. (2021). They developed their method using ERA-Interim datasets (European Centre for Medium-Range Weather Forecasts, 2006) and then applied it to super-resolution precipitation in Japan's Kuma River Watershed.

Temporal downscaling in the literature is not as profound as spatial downscaling. Lee and Jeong (2014) employed a stochastic take with the genetic algorithm in order to increase the temporal resolution of daily precipitation into hourly. Downscaling daily rain gauge measured precipitation into sub-daily precipitation data with six-hours of temporal resolution, Ryo et al. (2014) used satellite rainfall products. Seo and Krajewski (2015) increased the temporal resolution of rain maps produced by a QPE system from 5 minutes to 1 minute using advection correction and validated the results with measurements of ground rain gauges. We are not aware of any other studies that increase the temporal resolution of 2D rainfall maps.

Aside from super-resolution, or downscaling, on the other hand, there are studies that impute missing values in rainfall sequences. In such a study, Norazizi and Deni (2019) utilized ANNs to complete rainfall time series for the data from the Malaysian Meteorological Department. Similarly, for a spatiotemporal imputation of monthly rainfall data in 45 gauge stations located in southwestern Colombia, Canchala-Nastar et al. (2019) trained an ANN. In a similar manner to this study, Gao et al. (2021) tested two deep ANNs for temporal imputation of 2D rainfall maps. They comparatively employed a 3DCNN and a Convolutional Bidirectional LSTM and showed that the LSTM-based method they proposed outperforms the 3DCNN as well as some baselines, namely, optical flow, linear interpolation, and nearest frame.

Since the approach this paper utilizes is similar to video frame interpolation, computer vision literature provides more insight regarding the problem and plausible ways to address it. Besides the widely established method of calculating optical flow with statistical methods in order to generate videos with better frame rates, there are many studies exploring deep neural networks. Niklaus et al., 2017a and 2017b presented two incremental methods for converting low-temporal-resolution videos to high-temporal-resolution videos that used adaptive convolutions.

In another study, Liu et al. (2017) combined traditional optical flow calculation methods with deep neural networks in a method called deep voxel flow, to both temporally interpolate and extrapolate videos. Beyond that, Jiang et al. (2018) used a neural network structure (Unet) that was built for image segmentation purposes and trained it to learn optical flow to generate slow-motion videos out of regular low frame-rate videos. In a similar but extensive effort, Xiang et al. (2020) proposed a Convolutional LSTM-based model for both spatial and temporal interpolation of video frames. Cheng and Chen (2021), using separable convolutions, proposed an approach, namely enhanced deformable separable convolution, that aimed to provide a way to increase the frame rates of videos by estimating more than a frame between two available frames.

## 3. Methodology

This section describes the methods that we have employed in this paper. We start by describing the dataset used, IowaRain (Sit et al., 2021b), in the following subsection. We, then, provide details about the nearest frame, the optical flow and the CNN baselines as well as the CNN-based TempNet neural network architecture proposed in the study. Where, the baseline and TempNet models will produce a rain map at t=5, $t_5$ for given consecutive maps at t=0, $t_0$ and t=10, $t_{10}$. Finally, we describe how the neural networks were trained in the study.

### 3.1. Dataset

The dataset used in this paper is a rainfall event dataset, namely, IowaRain. The IowaRain dataset mainly relies on the Iowa Flood Center's QPE system (Seo and Krajewski 2020) using seven NEXRAD radars to cover the entire state of Iowa (Figure 1). The data IowaRain provides five minute and approximately 500m resolutions in time and space, respectively, and actively used in flood forecasting and mapping studies (Hu and Demir, 2021; Li et al., 2022). IowaRain consists of 288 rainfall events from 2016 to 2019. Each rainfall event is formed by a set of temporally consecutive 2D rain maps or snapshots for each timestamp with a five-minute interval.

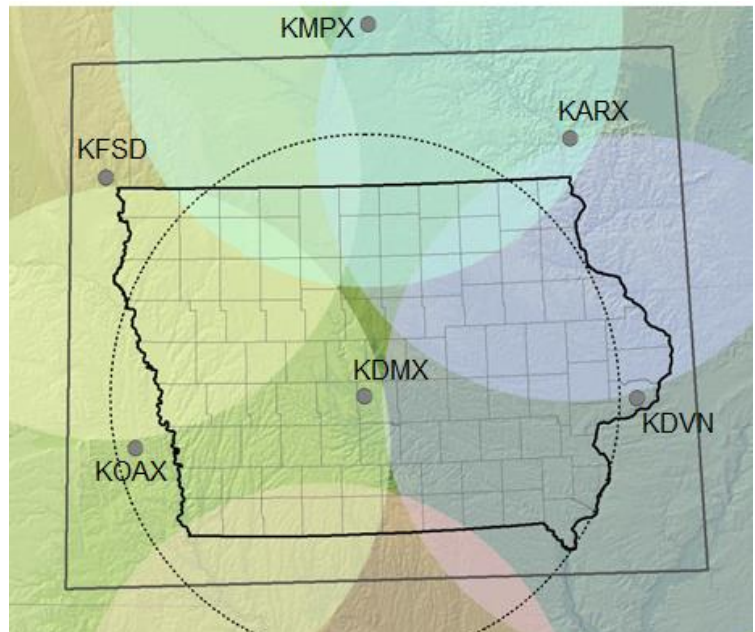

**Figure 1.** Locations of NEXRAD radars used in IowaRain and the boundary box showing the coverage over and around the state of Iowa

To prepare the dataset for training and testing, we formed a dataset entry for each snapshot $t_s$ in a rain event that has another snapshot coming right after ($t_{s+5}$) and right before ($t_{s-5}$) it. Each dataset entry consists of $t_{s-5}$ and $t_{s+5}$ as the input and $t_s$ as the output (Figure 2). For instance, a sequence of snapshots with a length of 10 would yield eight dataset entries. Then, in order to augment the dataset for better training, we form additional dataset entries by reversing the order of snapshots in the event and, by doing so, we double the number of dataset entries for the training dataset.

The IowaRain dataset chronologically provides 64, 67, 76, and 81 events for each year between 2016 and 2019. In order to do the dataset separation in a fashion closer to a 70/30 split (%70 training, %30 testing), we decided to use the rainfall events for 2019 as the test set and all the rainfall events before 2019 as the training set, making our set lengths 207 and 81 for the training and test sets, respectively. The final dataset entries sum up to 35,258 and 6,725 2D rainfall maps for the training and testing sets, respectively. It should be noted that aside from the test/train split and augmenting the dataset by reversing frame sequences, there was no form of preprocessing. Since the min-max normalization was already applied on the dataset, we did not make any further modifications to the measurements.

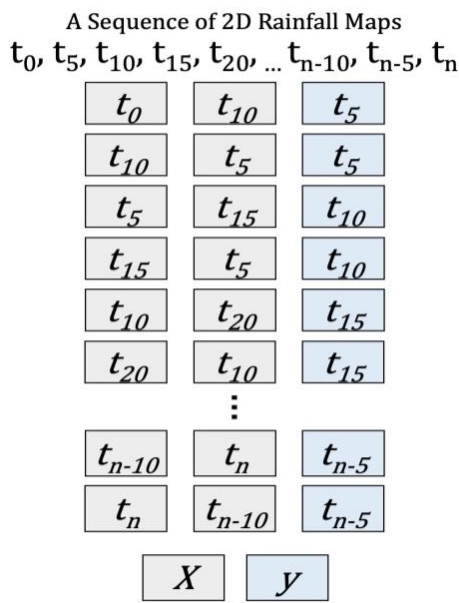

**Figure 2.** An example sequence of 2D rainfall maps and how a set of input and output pairs are built from that sequence

### 3.2. Nearest Frame

As one of the baselines, we employed the nearest frame as it was previously used in radar rainfall imputation comparisons (Gao et al., 2021). The nearest frame simply refers to assuming the interpolated frame is the same as the temporally closest frame to the forecasted frame. In other words when interpolating $t_5$ from $t_0$ and $t_{15}$, $t_5$ is assumed to be the same as $t_0$. In our case of interpolating $t_5$ from $t_0$ and $t_{10}$, on the other hand, we will use the predecessor frame, $t_0$, even though both input frames have the same distance to the output frame.

### 3.3. Optical Flow

In order to form a baseline method that is comparable to the model we propose, we chose to use an optical flow-based temporal interpolation method. This method depends on the optical flow calculation between two 2D rain maps and creating an intermediate frame between those rain maps that is at the same distance temporally. Optical flow has been widely used in the radar rainfall literature (Gao et al., 2021) as a method to build deep learning forecasts upon (Yan et al., 2021; Nie et al., 2021), as a baseline method to compare in temporal extrapolation of radar rainfall products (Kumar et al., 2020; Xie et al., 2020; Ayzel et al., 2020; Xu et al., 2021) as well as in temporal imputation of 2D rainfall maps (Gao et al., 2021) since it was first proposed by Bowler et al. (2004).

Optical flow is the summary of the spatial changes of objects between two frames of a scene. In other words, it shows the motion of the objects in order to estimate the mobility of velocity fields between two frames. Consequently, an optical flow calculation algorithm is a way to determine how objects move from one scene to another. What an optical flow calculation algorithm actually does is to calculate the change of pixel intensities. In computer vision, *pixel intensity* is defined as the properties a pixel carries in a scene. For a 2D tensor, each value's pixel intensity is determined by its position in the 2D tensor and its value. The same phenomenon is also valid for the case we tackle in this study since rainfall maps are, in fact, 2D tensors.

There are many optical flow calculation algorithms in the computer vision literature. In this paper, we employed the Gunnar-Farneback optical flow (Farnebäck, 2003), or sometimes referred to as dense optical flow. Gunnar-Farneback optical flow calculates pixel intensities for each pixel in the scene, as opposed to feature-level intensity calculations, hence the name "dense", as opposed to sparse. For the rain map case, that would mean calculating the changes for each of the measurements for the 0.5km x 0.5km areas that make up the 2D rain map. The mathematical formulation of the algorithm is beyond the scope of this paper; we suggest the reader refer to the paper for further details. Once the optical flow is calculated, the next step in the approach is to transfer every measurement depending on their motion vectors in the calculated optical flow and their location in the first and second frame.

For comparison purposes, the optical flow between input snapshots was calculated for each of the entries in the test dataset. Then, color propagation was done using that calculation in order to estimate the intermediate frame. Once all the estimations were done for the testing dataset, the performance metrics reported in the next section were calculated using estimated and actual snapshots. The entirety of the baseline method was implemented using the Gunnar-Farneback optical flow implementation using the OpenCV (Bradski & Kaehler, 2000) and NumPy (Harris et al., 2020) numeric computing libraries.

### 3.4. Convolutional Neural Networks (CNNs)

Being primarily used in computer vision tasks, Convolutional Neural Networks, ConvNets, or simply CNNs, work relying on convolutional layers. A convolutional layer in a neural network uses weight tensors called filters to explore patterns in inputs. In contrast to weights in a fully connected neural network layer, a filter in a convolutional layer typically works two-dimensionally in order to grasp multi-dimensional (both vertical and horizontal in a 2D black and white image, or in a 2D rain map, similarly) spatial relations over the given input tensor. Building upon this core idea, a CNN with multiple convolutional layers can map complex spatial relations in a given tensor that only one convolutional layer fails to achieve (Goodfellow et al., 2016). For further technical details regarding CNNs, readers could refer to LeCun (1989) and Goodfellow et al. (2016).

Unlike typical artificial neural networks, CNNs work 2-dimensionally over a tensor. Since 2D rainfall maps fall into that description, excluding their temporal dimension, one can assume CNNs are the go-to neural network structures for interpolation of 2D rainfall maps. Since CNNs are typically used with images containing three color channels, a CNN that processes an input image typically has three filters. The dataset we employed, on the other hand, by its nature, has only one channel, a rain rate (mm/h) map.

**Table 1. Architecture details for the CNN-baseline.**

| Layer | Input Channels | Output Channels | Details |
|---|---|---|---|
| conv_1 | 2 | 3 | kernel=3, stride=1, padding=1 |
| conv_2 | 3 | 5 | kernel=3, stride=1, padding=1 |
| conv_3 | 5 | 3 | kernel=3, stride=1, padding=1 |
| conv_4 | 3 | 1 | kernel=3, stride=1, padding=1 |

To form a CNN baseline that can be compared to TempNet, we built a 4-layer CNN where Rectified Linear Unit (ReLU) was used as the activation function after each layer, namely CNN-baseline. The input frames for the CNN were concatenated into a 3D tensor, as if they were separate channels. In other words, when interpolating $t_5$, $t_0$ of shape 1088 x 1760 and $t_{10}$ of shape 1088 x 1760 were concatenated to form the input tensor of shape 2 x 1088 x 1760. Details regarding the architecture can be found in Table 1.

### 3.5. TempNet

In order to explore how the temporal resolution of 2D rain maps could be improved, we propose a CNN-based neural network architecture, TempNet. The TempNet provides a fast but effective alternative to the optical flow-based color transfer method and builds upon the CNN architecture we have described in previous subsections.

In a different manner and scale than the CNN-baseline, a CNN that works on a single channel of input was built. However, since the neural network expects two 2D rain maps as input, we had to use two different series of convolutional layers to learn the pattern over them first. Then the difference between those convolutional layers' outputs was added to the first frame in a skip connection fashioned architectural design choice that was introduced to improve the convergence of neural networks over the training data (He et al., 2016). In the end, that summation was fed to another series of convolutional layers to output the intermediate frame between two input frames (Figure 3).

Both CNN-baseline and the TempNet described here were trained using L1 Loss (also known as Mean Absolute Error - MAE) as the cost function and Adam optimizer (Kingma and Ba, 2014) with the help of a Reduce-on-Plateau learning rate scheduler over the training loss. The network was trained on the training dataset described in the Dataset subsection on NVIDIA Titan V

GPUs using the PyTorch numeric computing library (Paszke et al., 2019). As for hyperparameters, a batch size of 32, an initial learning rate of 1e-3, and a number of epochs of 50 without early stop were used.

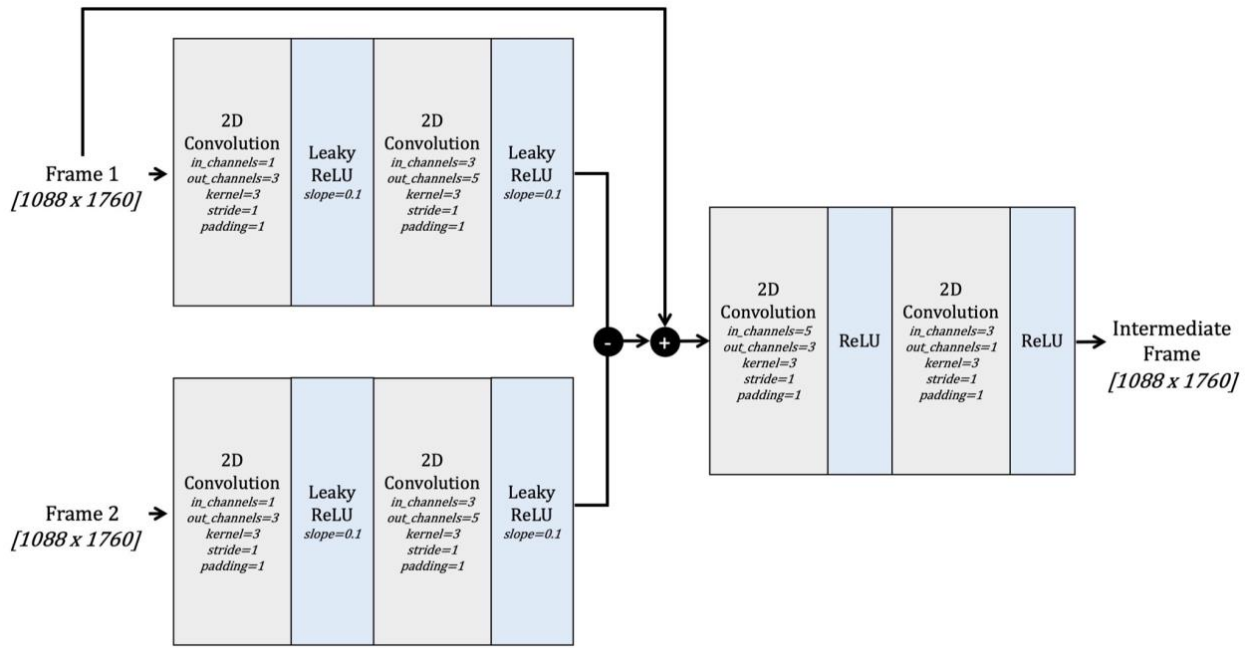

**Figure 3.** Architecture scheme for the TempNet model

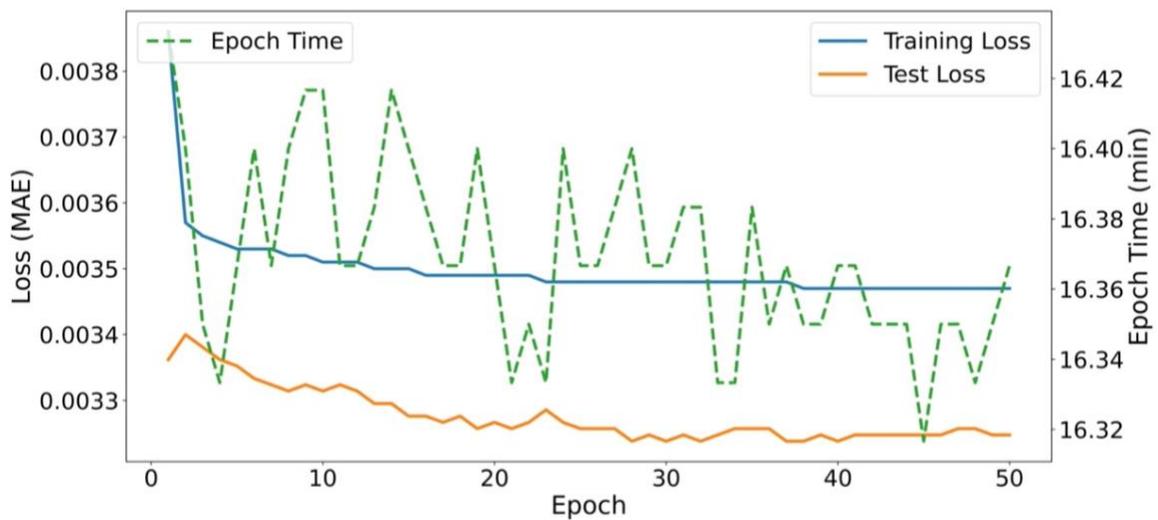

**Figure 4.** Change in epoch time, training loss, and test loss over 50 epochs of training for the CNN-baseline

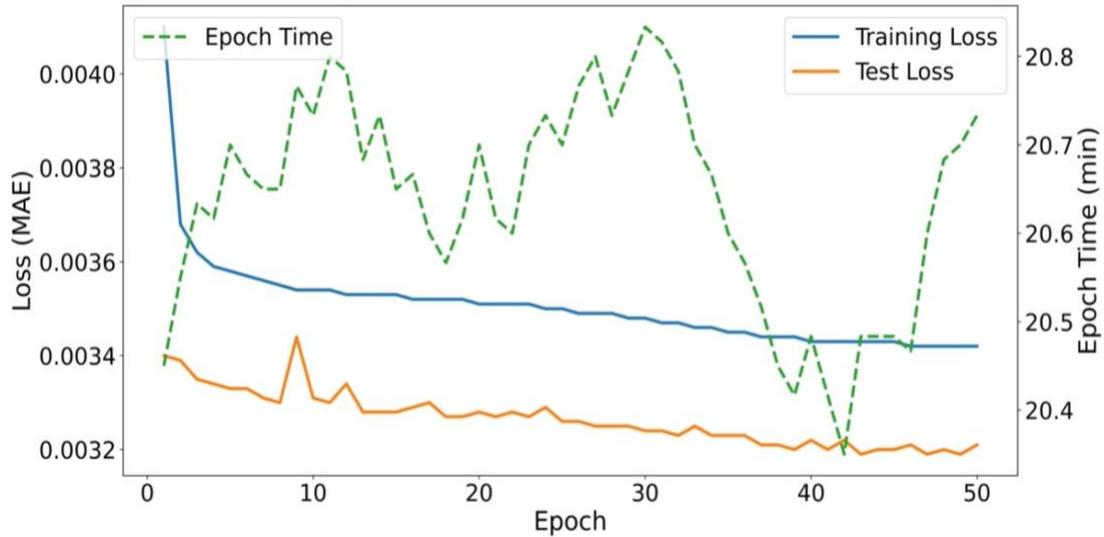

**Figure 5.** Change in epoch time, training loss, and test loss over 50 epochs of training for the TempNet

## 4. Results and Discussion

This section defines performance metrics and presents results using those metrics for nearest frame, optical flow, CNN-baseline, and the TempNet. The loss changes (L1 Loss) over both training and testing datasets during 50 epochs of training are given in Figures 4 and 5 for the CNN-baseline and the TempNet, respectively. As the figures suggest, the performances of both networks steadily increase over epochs and stabilizes after the 40th epoch. It is worth noting that the costs reported in Figures 4 and 5 are averaged over the respective datasets, and they are for normalized datasets.

**Table 2.** Contingency table for Hit (H), Miss (M) and False Alarm (F)

|  |  | **Real Values** | |
|---|---|---|---|
|  |  | **Rain** | **No Rain** |
| **Predicted Values** | **Rain** | H | F |
|  | **No Rain** | M | - |

We reported four metrics for each of the methods described earlier, namely Mean Absolute Error (MAE), Probability of Detection (POD), False Alarm Ratio (FAR), and Critical Success Index (CSI). MAE reports the mean of the absolute values of the differences between estimated and actual 2D rain maps in the test dataset. POD, FAR, and CSI are calculated by (1–3), respectively, using the number of hits (H), false alarms (F), and mises (M). In this case, H, F, and M are calculated in binary fashion (e.g., threshold is zero); H is defined as the number of correctly estimated rainfall cells, or in other words, the number of elements in the 2D tensor that were correctly estimated as non-zero values. F is defined as the number of wrongly estimated rainfall cells: while the cells were estimated to have rain, the ground truth for the same indices in the 2D map had zero. M is defined as the number of rainfall cells that were estimated as zero, while they have a non-zero value in ground truth. It should be noted that while the best value for POD and CSI is 1.0 (the greater the better), it is 0.0 for FAR (the lower the better).

$$POD = \frac{H}{H + M} \quad \text{(Eq. 1)}$$

$$FAR = \frac{F}{H + F} \quad \text{(Eq. 2)}$$

$$CSI = \frac{H}{H + F + M} \quad \text{(Eq. 3)}$$

Table 3 summarizes the performance of best CNN-baseline and TempNet models for training over the testing dataset as well as the nearest frame and optical flow's performance using the metrics mentioned above in interpolating $t_5$ from $t_0$ and $t_{10}$. As Table 3 suggests, both TempNet and CNN-baseline outperform the nearest frame approach and the optical flow-based interpolation baseline method for the MAE metric, over which neural network approaches were trained. Among the two neural network models, TempNet provides slightly better results at the expense of a slightly longer training time, as Figures 4 and 5 depict. Thus, for a more accurate temporal resolution increase, while TempNet offers a better solution than optical flow and a basic CNN baseline, the CNN-baseline also presents good enough results in terms of MAE.

Beyond accuracy, one upside of using neural network models is the runtime. Since calculating optical flow and building the new frame out of two sequential frames needs work over individual pixels or measurements, parallelization is challenging and, consequently, time-consuming. Conversely, the same task takes a trivial amount of time on both GPUs and CPUs compared to the baseline method's runtime using TempNet or CNN-baseline once the training is done. As for POD, the optical flow approach seems to be a better methodology, but coupled with the FAR, one can easily suggest that optical flow fills radically more cells with values than other approaches. In other words, optical flow focuses on true positives while estimating significantly more false positives than either of the neural network models estimate. Another take here for neural networks is that optical flow being better at POD is most likely due to the fact that the neural networks we presented were trained by optimizing them over MAE. One can argue that optimizing networks' weights by other metrics would make them better than optical flow as well.

**Table 3.** Performance summary of tested methods for predicting intermediate frame

| Methodology | MAE (mm/h) ▼ | FAR ▼ | CSI ▲ | POD ▲ |
|---|---|---|---|---|
| **Nearest Frame** | 0.533 | 0.102 | 0.841 | 0.922 |
| **Optical Flow** | 0.35 | 0.151 | 0.832 | **0.975** |
| **CNN-baseline** | 0.341 | 0.074 | **0.865** | 0.928 |
| **TempNet** | **0.332** | **0.073** | 0.864 | 0.925 |

Applying a single frame temporal interpolation method recursively to generate several intermediate frames is an interesting idea, but there are some drawbacks to this approach. First, recursive temporal interpolation cannot be fully parallelized because certain frames cannot be computed until others are generated. Second, only $2i-1$ intermediate frames can be generated (e.g., 3, 5, 7). Third, during recursive interpolation, errors also build up. Subsequently, in order to understand how recursion affects the performance of the approaches presented, we have run a series of tests. Our first test is to understand how the baseline methods and TempNet perform in interpolating $t_{10}$ from $t_0$ and $t_{20}$, namely estimating $t'_{10}$. Table 4 shows each method's performance on this task using the aforementioned metrics. According to the presented scores, the performances of the models change drastically in recursion, and CNN-baseline and TempNet lose first place to optical flow. Furthermore, the change in performance suggests that any of the presented methodologies don't scale well enough. Nevertheless, considering optical flow calculates the difference between two frames and fills in the middle frame by moving the pixels by half of the calculated motion vectors, and TempNet and CNN-baseline performed similarly to optical flow in terms of MAE while providing better FAR and CSI scores, we can infer that neural networks are promising in interpolating the temporal distances they were not trained for. It also deserves mentioning that optical flow seems to spatially over-estimate rainfall again since FAR is significantly greater than that of other approaches.

**Table 4.** Performance summary of methodologies in interpolating frames 10-minutes apart.

| Methodology | MAE (mm/h) ▼ | FAR ▼ | CSI ▲ | POD ▲ |
|---|---|---|---|---|
| **Nearest Frame** | 0.728 | 0.152 | 0.739 | 0.846 |
| **Optical Flow** | **0.499** | 0.208 | 0.768 | **0.961** |
| **CNN-baseline** | 0.506 | 0.136 | **0.788** | 0.896 |
| **TempNet** | 0.503 | **0.129** | **0.788** | 0.888 |

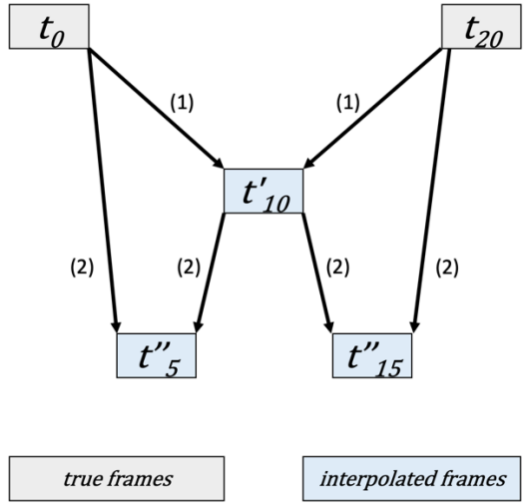

**Figure 6.** Workflow of iterative estimation of t''$_5$ and t''$_{15}$

**Table 5.** Performances of methodologies in interpolating frames in second iteration.

| Methodology | MAE (mm/h) ▼ | FAR ▼ | CSI ▲ | POD ▲ |
|---|---|---|---|---|
| Nearest Frame | 0.686 | 0.144 | 0.753 | 0.854 |
| Optical Flow | 0.593 | 0.215 | 0.757 | **0.952** |
| CNN-baseline | **0.422** | **0.088** | **0.82** | 0.888 |
| TempNet | **0.422** | 0.089 | 0.819 | 0.888 |

Furthermore, to see how errors build up in recursion, we performed a test where an already interpolated frame of t'$_{10}$ from t$_0$ and t$_{20}$ was used. In other words, frames t''$_5$ and t''$_{15}$ were interpolated using t$_0$ and t'$_{10}$, and t'$_{10}$ and t$_{20}$ respectively (Figure 6). Table 5 shows the performance of each of the proposed methods in iterative estimation. While neural network models present the best results in terms of the average absolute error, they are still not as good as the optical flow approach in providing a solution when it comes to POD. Nevertheless, the fact that optical flow is drastically worse at FAR changes the tie in favor of neural network-based methods.

Figures 7 and 8 show ground truth and each methodology's output in interpolating frames 5 minutes apart by using the color scheme of the Iowa Flood Information System (Demir et al., 2018). According to the rainfall maps depicted in these figures, we can easily see that radar rainfall maps have small areas of precipitation scattered around larger areas that are denser. This phenomenon could be because of the artifacts caused by radars or the QPE process. Optical flow seems to be transferring those artifact-like small regions relatively better than other approaches. However, it also distributes the artifacts as well as larger regions and forms shadows of small precipitation regions all around the map. Coupled with Table 3, optical flow's ability to transfer small areas of precipitation makes them in-overall better at POD, and creating those shadows makes them estimate false positives, thus resulting in a worse FAR score. On the other hand, performing similarly, CNN-baseline and TempNet get rid of those artifact-like small regions by producing relatively blurry outputs. This tendency to ignore small areas to focus on larger areas comes with an expense. More often than not, intense rainfall happens in smaller areas, and since the CNN-based approaches we presented tend to ignore changes in small areas, they are also inclined to not estimate intense rainfall as accurately as optical flow.

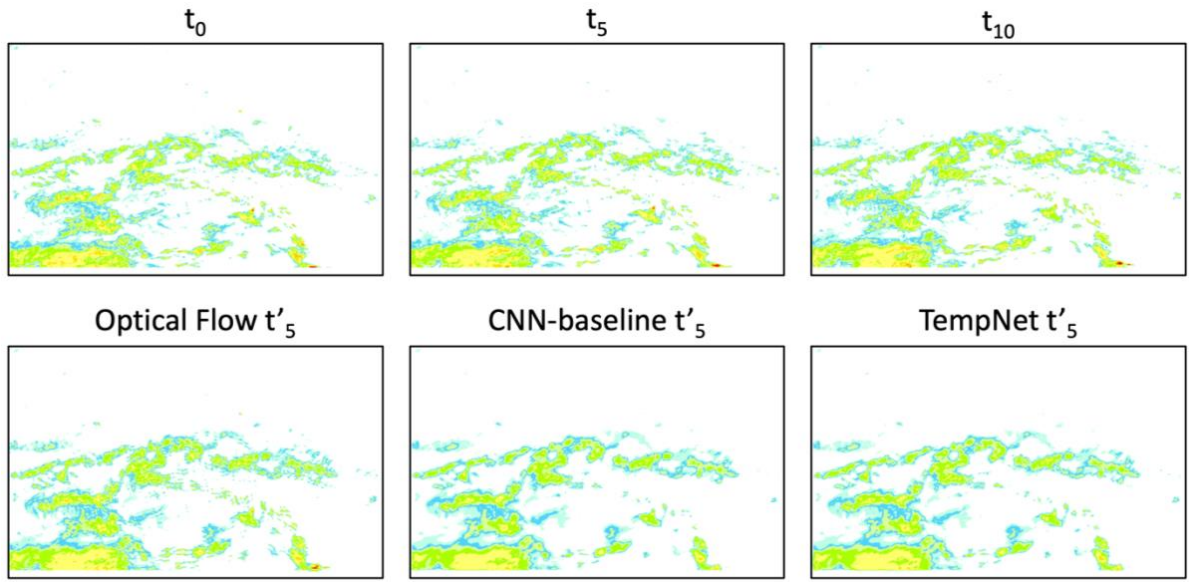

**Figure 7.** An example 10-minutes sequence of ground truth and estimations by optical flow, CNN-baseline and TempNet from a rainfall event on May 8th, 2019

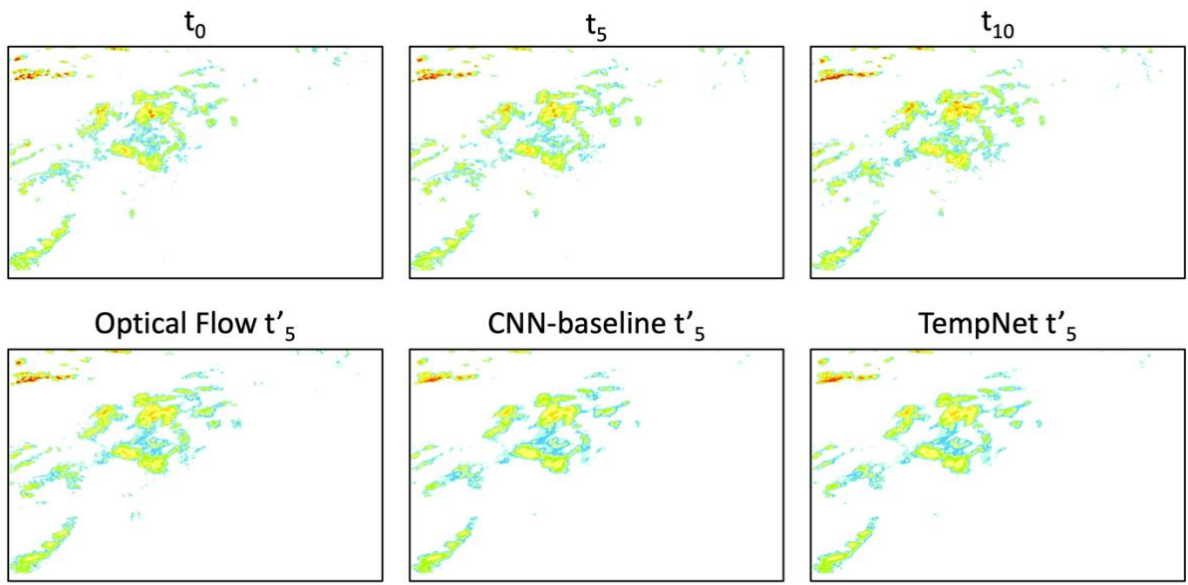

**Figure 8.** An example 10-minutes sequence of ground truth and estimations by optical flow, CNN-baseline and TempNet from a rainfall event on June 11th, 2019

Finally, for comparison purposes, we did some tests to understand the visual performance of the CNN-baseline and the TempNet by increasing the temporal resolution of some events in the test dataset by three iterations over the resolutions they were not trained for. In other words, rainfall maps' temporal resolution was increased from 5-minutes to 0.7-minutes. After visualizing the actual and generated rain maps with the Iowa Flood Information System's rainfall map color scheme, even though we were able to see that the TempNet and the CNN-baseline were not able to carry all the information regarding the rainfall in specific areas, they were still able to represent most of the accumulated rainfall clearly. Alternatively, the optical flow method would visually average two input frames that had lots of shadows scattered over the maps.

## 5. Conclusions

In this study, we presented a CNN-based temporal resolution improvement model, TempNet, and compared it to the nearest frame approach, an optical flow-based baseline method, and a CNN-based baseline method, namely CNN-baseline. Results show that TempNet broadly outperforms baseline models according to MAE as well as a combination of other metrics such as False Alarm Ratio, Probability of Detection, and Critical Success Index, while doing that with a significantly better computation time than optical flow. We consider this work as a significant step towards creating better rainfall maps for hydrological modeling needs such as flood forecasting (Sit et al., 2021a; Xiang et al., 2021), climate change modeling (Rolnick et al., 2019; Sit et al., 2020) and missing rainfall map imputation (Gao et al., 2021). As for future directions of this research, we aim to extend this study for better interpolation accuracy as well as faster computation times by building more efficient neural networks to support operational flood forecasting needs (Xiang and Demir, 2022b). In order to train better models, various loss functions could be utilized, such as a loss function where the metrics we used in this are combined by differentiable weights.

One challenge with the process was that weather radar-based rainfall maps contain significant artifacts. Although the IowaRain dataset provides rain maps that are vastly cleansed of those artifacts, some remaining noise (e.g., ground and wind turbine clutter; Seo et al., 2015) may affect both accuracy and visual representations. In order to test the dataset better and to develop more accurate models for both imputation and temporal super-resolution of radar rainfall products, datasets with fewer artifacts would become useful. Thus, in order to create improved datasets, as a future perspective, we aim to focus on denoising radar rainfall maps to decrease the number of artifacts in the original dataset with neural networks and train the TempNet over a cleaner dataset. This highlights the need for better benchmark datasets for research communities (Demir et al., 2022). Another future aspect for this problem would be to rectify the issues we mentioned regarding recursive temporal interpolation by introducing a neural network that is capable of producing intermediate frames for any given temporal distance to the input frames, as in Jiang et al. (2018).

## References

1. Alabbad, Y., Yildirim, E. and Demir, I., 2022. Flood mitigation data analytics and decision support framework: Iowa Middle Cedar Watershed case study. Science of The Total Environment, 814, p.152768.

2. Alizamir, M., Azhdary Moghadam, M., Hashemi Monfared, A. and Shamsipour, A., 2017. A hybrid artificial neural network and particle swarm optimization algorithm for statistical downscaling of precipitation in arid region. *Ecopersia*, *5*(4), pp.1991-2006.
3. Anagnostou, E.N. and Krajewski, W.F., 1999. Real-time radar rainfall estimation. Part I: Algorithm formulation. *Journal of Atmospheric and Oceanic Technology*, *16*(2), pp.189-197.
4. Atencia, A., Mediero, L., Llasat, M.C. and Garrote, L., 2011. Effect of radar rainfall time resolution on the predictive capability of a distributed hydrologic model. *Hydrology and Earth System Sciences*, *15*(12), pp.3809-3827.
5. Ayzel, G., Scheffer, T. and Heistermann, M., 2020. RainNet v1. 0: a convolutional neural network for radar-based precipitation nowcasting. *Geoscientific Model Development*, *13*(6), pp.2631-2644.
6. Bowler, N.E., Pierce, C.E. and Seed, A., 2004. Development of a precipitation nowcasting algorithm based upon optical flow techniques. *Journal of Hydrology*, *288*(1-2), pp.74-91.
7. Bradski, G. and Kaehler, A., 2000. OpenCV. *Dr. Dobb's journal of software tools*, *3*.
8. Cahoon, L.B. and Hanke, M.H., 2017. Rainfall effects on inflow and infiltration in wastewater treatment systems in a coastal plain region. *Water Science and Technology*, *75*(8), pp.1909-1921.
9. Canchala-Nastar, T., Carvajal-Escobar, Y., Alfonso-Morales, W., Cerón, W.L. and Caicedo, E., 2019. Estimation of missing data of monthly rainfall in southwestern Colombia using artificial neural networks. *Data in brief*, *26*, p.104517.
10. Cheng, X. and Chen, Z., 2021. Multiple video frame interpolation via enhanced deformable separable convolution. *IEEE Transactions on Pattern Analysis and Machine Intelligence*.
11. Demir, I., Jiang, F., Walker, R.V., Parker, A.K. and Beck, M.B., 2009, October. Information systems and social legitimacy scientific visualization of water quality. In 2009 IEEE International Conference on Systems, Man and Cybernetics (pp. 1067-1072). IEEE.
12. Demir, I., Yildirim, E., Sermet, Y. and Sit, M.A., 2018. FLOODSS: Iowa flood information system as a generalized flood cyberinfrastructure. *International journal of river basin management*, *16*(3), pp.393-400.
13. Demir, I., Xiang, Z., Demiray, B., & Sit, M. 2022. WaterBench: A Large-scale Benchmark Dataset for Data-Driven Streamflow Forecasting. Earth System Science Data Discussions, 1-19.
14. Demiray, B.Z., Sit, M. and Demir, I., 2021a. D-SRGAN: DEM super-resolution with generative adversarial networks. SN Computer Science, 2(1), pp.1-11.
15. Demiray, B.Z., Sit, M. and Demir, I., 2021b. DEM Super-Resolution with EfficientNetV2. *arXiv preprint arXiv:2109.09661*.
16. ERA-Interim, European Centre for Medium-Range Weather Forecasts, 2006. Available online: https://www.ecmwf.int/en/forecasts/datasets/reanalysis-datasets/era-interim.
17. Ewing, G. and Demir, I., 2021. An ethical decision-making framework with serious gaming: a smart water case study on flooding. Journal of Hydroinformatics, 23(3), pp.466-482.
18. Fabry, F., Bellon, A., Duncan, M.R. and Austin, G.L., 1994. High resolution rainfall measurements by radar for very small basins: the sampling problem reexamined. *Journal of Hydrology*, *161*(1-4), pp.415-428.
19. Farnebäck, G., 2003, June. Two-frame motion estimation based on polynomial expansion. In *Scandinavian conference on Image analysis* (pp. 363-370). Springer, Berlin, Heidelberg.

20. Gao, L., Zheng, Y., Wang, Y., Xia, J., Chen, X., Li, B., Luo, M. and Guo, Y., 2021. Reconstruction of Missing Data in Weather Radar Image Sequences Using Deep Neuron Networks. *Applied Sciences*, *11*(4), p.1491.
21. Gautam, A., Sit, M. and Demir, I., 2020. Realistic river image synthesis using deep generative adversarial networks. Front. Water 4:784441. doi:10.3389/frwa.2022.784441
22. Goodfellow, I., Bengio, Y. and Courville, A., 2016. *Deep learning*. MIT press.
23. Harris, C.R., Millman, K.J., van der Walt, S.J., Gommers, R., Virtanen, P., Cournapeau, D., Wieser, E., Taylor, J., Berg, S., Smith, N.J. and Kern, R., 2020. Array programming with NumPy. *Nature*, *585*(7825), pp.357-362.
24. He, K., Zhang, X., Ren, S. and Sun, J., 2016. Deep residual learning for image recognition. In *Proceedings of the IEEE conference on computer vision and pattern recognition* (pp. 770-778).
25. Hou, Y.K., Chen, H., Xu, C.Y., Chen, J. and Guo, S.L., 2017. Coupling a Markov chain and support vector machine for at-site downscaling of daily precipitation. *Journal of Hydrometeorology*, *18*(9), pp.2385-2406.
26. Hu, A. and Demir, I., 2021. Real-time flood mapping on client-side web systems using hand model. Hydrology, 8(2), p.65.
27. Hu, Y.F., Yin, F.K. and Zhang, W.M., 2021. Deep learning-based precipitation bias correction approach for Yin–He global spectral model. *Meteorological Applications*, *28*(5), p.e2032.
28. Jha, M.K., Gassman, P.W. and Arnold, J.G., 2007. Water quality modeling for the Raccoon River watershed using SWAT. *Transactions of the ASABE*, *50*(2), pp.479-493.
29. Jiang, H., Sun, D., Jampani, V., Yang, M.H., Learned-Miller, E. and Kautz, J., 2018. Super slomo: High quality estimation of multiple intermediate frames for video interpolation. In *Proceedings of the IEEE Conference on Computer Vision and Pattern Recognition* (pp. 9000-9008).
30. Kingma, D.P. and Ba, J., 2014. Adam: A method for stochastic optimization. *arXiv preprint arXiv:1412.6980*.
31. Krajewski, W.F. and Smith, J.A., 2002. Radar hydrology: rainfall estimation. *Advances in water resources*, *25*(8-12), pp.1387-1394.
32. Krajewski, W.F., Kruger, A., Singh, S., Seo, B.C. and Smith, J.A., 2013. Hydro-NEXRAD-2: Real-time access to customized radar-rainfall for hydrologic applications. *Journal of Hydroinformatics*, *15*(2), pp.580-590.
33. Kumar, A., Islam, T., Sekimoto, Y., Mattmann, C. and Wilson, B., 2020. Convcast: An embedded convolutional LSTM based architecture for precipitation nowcasting using satellite data. *Plos one*, *15*(3), p.e0230114.
34. LeCun, Y., 1989. Generalization and network design strategies. *Connectionism in perspective*, *19*, pp.143-155.
35. Lepetit, P., Ly, C., Barthès, L., Mallet, C., Viltard, N., Lemaitre, Y. and Rottner, L., 2021. Using deep learning for restoration of precipitation echoes in radar data. *IEEE Transactions on Geoscience and Remote Sensing*, *60*, pp.1-14.
36. Li, Z., Mount, J. and Demir, I., 2022. Accounting for uncertainty in real-time flood inundation mapping using HAND model: Iowa case study. Natural Hazards, 112(1), pp.977-1004.

37. Li, W., Zhou, W., Wang, Y.M., Shen, C., Zhang, X. and Li, X., 2019, December. Meteorological radar fault diagnosis based on deep learning. In *2019 International Conference on Meteorology Observations (ICMO)* (pp. 1-4). IEEE.
38. Liu, C. and Krajewski, W.F., 1996. A Comparison of Methods for Calculation of Radar-Rainfall Hourly Accumulations 1. *JAWRA Journal of the American Water Resources Association*, *32*(2), pp.305-315
39. Liu, Z., Yeh, R.A., Tang, X., Liu, Y. and Agarwala, A., 2017. Video frame synthesis using deep voxel flow. In *Proceedings of the IEEE International Conference on Computer Vision* (pp. 4463-4471).
40. Muste, M., Lyn, D.A., Admiraal, D., Ettema, R., Nikora, V. and García, M.H. eds., 2017. Experimental Hydraulics: Methods, Instrumentation, Data Processing and Management: Volume I: Fundamentals and Methods. CRC Press.
41. Nie, T., Ji, X. and Pang, Y., 2021, December. OFAF-ConvLSTM: An Optical Flow Attention Fusion-ConvLSTM Model for Precipitation Nowcasting. In *2021 3rd International Academic Exchange Conference on Science and Technology Innovation (IAECST)* (pp. 283-286). IEEE.
42. Niklaus, S., Mai, L. and Liu, F., 2017. Video frame interpolation via adaptive convolution. In *Proceedings of the IEEE Conference on Computer Vision and Pattern Recognition* (pp. 670-679).
43. Niklaus, S., Mai, L. and Liu, F., 2017. Video frame interpolation via adaptive separable convolution. In *Proceedings of the IEEE International Conference on Computer Vision* (pp. 261-270).
44. Paszke, A., Gross, S., Massa, F., Lerer, A., Bradbury, J., Chanan, G., Killeen, T., Lin, Z., Gimelshein, N., Antiga, L. and Desmaison, A., 2019. Pytorch: An imperative style, high-performance deep learning library. *Advances in neural information processing systems*, *32*, pp.8026-8037.
45. Pathak, C.S. and Vieux, B.E., 2008. Improvement of NEXRAD rainfall data for Central and South Florida. In *World Environmental and Water Resources Congress 2008: Ahupua'A* (pp. 1-16).
46. Rolnick, D., Donti, P.L., Kaack, L.H., Kochanski, K., Lacoste, A., Sankaran, K., Ross, A.S., Milojevic-Dupont, N., Jaques, N., Waldman-Brown, A. and Luccioni, A., 2019. Tackling climate change with machine learning. *arXiv preprint arXiv:1906.05433*.
47. Ryo, M., Saavedra Valeriano, O.C., Kanae, S. and Ngoc, T.D., 2014. Temporal downscaling of daily gauged precipitation by application of a satellite product for flood simulation in a poorly gauged basin and its evaluation with multiple regression analysis. *Journal of Hydrometeorology*, *15*(2), pp.563-580.
48. Salimi, A.H., Masoompour Samakosh, J., Sharifi, E., Hassanvand, M.R., Noori, A. and von Rautenkranz, H., 2019. Optimized artificial neural networks-based methods for statistical downscaling of gridded precipitation data. *Water*, *11*(8), p.1653.
49. Seo, B.C. and Krajewski, W.F., 2015. Correcting temporal sampling error in radar-rainfall: Effect of advection parameters and rain storm characteristics on the correction accuracy. *Journal of Hydrology*, *531*, pp.272-283.

50. Seo, B.C. and Krajewski, W.F., 2020. Statewide real-time quantitative precipitation estimation using weather radar and NWP model analysis: Algorithm description and product evaluation. *Environmental Modelling & Software*, *132*, p.104791.
51. Seo, B.C., Krajewski, W.F. and Mishra, K.V., 2015. Using the new dual-polarimetric capability of WSR-88D to eliminate anomalous propagation and wind turbine effects in radar-rainfall. *Atmospheric Research*, *153*, pp.296-309.
52. Seo, B.C., Krajewski, W.F., Quintero, F., Buan, S. and Connelly, B., 2021. Assessment of Streamflow Predictions Generated Using Multimodel and Multiprecipitation Product Forcing. *Journal of Hydrometeorology*, *22*(9), pp.2275-2290.
53. Seo, B.C., Quintero, F. and Krajewski, W.F., 2018. High-resolution QPF uncertainty and its implications for flood prediction: A case study for the eastern Iowa flood of 2016. *Journal of Hydrometeorology*, *19*(8), pp.1289-1304.
54. Sit, M., Demiray, B.Z., Xiang, Z., Ewing, G.J., Sermet, Y. and Demir, I., 2020. A comprehensive review of deep learning applications in hydrology and water resources. *Water Science and Technology*, *82*(12), pp.2635-2670.
55. Sit, M., Demiray, B. and Demir, I., 2021a. Short-term Hourly Streamflow Prediction with Graph Convolutional GRU Networks. *arXiv preprint arXiv:2107.07039*.
56. Sit, M., Seo, B.C. and Demir, I., 2021b. IowaRain: A Statewide Rain Event Dataset Based on Weather Radars and Quantitative Precipitation Estimation. *arXiv preprint arXiv:2107.03432*.
57. Shapiro, A., Willingham, K.M. and Potvin, C.K., 2010. Spatially variable advection correction of radar data. Part I: Theoretical considerations. *Journal of the Atmospheric Sciences*, *67*(11), pp.3445-3456.
58. Teague, A., Y. Sermet, I. Demir, and M. Muste. "A collaborative serious game for water resources planning and hazard mitigation." International Journal of Disaster Risk Reduction 53 (2021): 101977.
59. Tu, T., Ishida, K., Ercan, A., Kiyama, M., Amagasaki, M. and Zhao, T., 2021. Hybrid precipitation downscaling over coastal watersheds in Japan using WRF and CNN. *Journal of Hydrology: Regional Studies*, *37*, p.100921.
60. Villarini, G. and Krajewski, W.F., 2010. Review of the different sources of uncertainty in single polarization radar-based estimates of rainfall. *Surveys in geophysics*, *31*(1), pp.107-129.
61. Xiang, X., Tian, Y., Zhang, Y., Fu, Y., Allebach, J.P. and Xu, C., 2020. Zooming slow-mo: Fast and accurate one-stage space-time video super-resolution. In *Proceedings of the IEEE/CVF conference on computer vision and pattern recognition* (pp. 3370-3379).
62. Xiang, Z., Demir, I., Mantilla, R. and Krajewski, W.F., 2021. A Regional Semi-Distributed Streamflow Model Using Deep Learning. EarthArxiv, 2152. doi.org/10.31223/X5GW3V
63. Xiang, Z. and Demir, I., 2022a. Fully distributed rainfall-runoff modeling using spatial-temporal graph neural network. EarthArxiv, 3018. [doi.org/10.31223/X57P74](doi.org/10.31223/X57P74)
64. Xiang, Z. and Demir, I., 2022b. Real-Time Streamflow Forecasting Framework, Implementation and Post-Analysis Using Deep Learning. EarthArxiv, 3162. [https://doi.org/10.31223/X5BW6R](https://doi.org/10.31223/X5BW6R)

65. Xie, P., Li, X., Ji, X., Chen, X., Chen, Y., Liu, J. and Ye, Y., 2020. An energy-based generative adversarial forecaster for radar echo map extrapolation. *IEEE Geoscience and Remote Sensing Letters*.
66. Xu, F., Li, G., Du, Y., Chen, Z. and Lu, Y., 2021, May. Multi-Layer Networks for Ensemble Precipitation Forecasts Postprocessing. In *Proceedings of the AAAI Conference on Artificial Intelligence* (Vol. 35, No. 17, pp. 14966-14973).
67. Yan, B.Y., Yang, C., Chen, F., Takeda, K. and Wang, C., 2021. FDNet: A Deep Learning Approach with Two Parallel Cross Encoding Pathways for Precipitation Nowcasting. *arXiv preprint arXiv:2105.02585*.
68. Yildirim, E. and Demir, I., 2021. An integrated flood risk assessment and mitigation framework: A case study for middle Cedar River Basin, Iowa, US. International Journal of Disaster Risk Reduction, 56, p.102113.
69. Zhang, J., Howard, K., Langston, C., Vasiloff, S., Kaney, B., Arthur, A., Van Cooten, S., Kelleher, K., Kitzmiller, D., Ding, F. and Seo, D.J., 2011. National Mosaic and Multi-Sensor QPE (NMQ) system: Description, results, and future plans. *Bulletin of the American Meteorological Society*, *92*(10), pp.1321-1338.